\documentclass[runningheads]{llncs}

\usepackage{amsmath}
\usepackage{graphicx}
\usepackage{todonotes}
\usepackage{svg}
\usepackage{arydshln}
\usepackage{amsmath}
\usepackage{amssymb}
\usepackage{float}

\begin{document}
\title{Enhancing Weather Predictions: Super-Resolution via Deep Diffusion Models}
%
%
\author{Jan-Matyáš Martinů \inst{1} \and
Petr Šimánek\inst{1}\orcidID{0000-0001-5808-0865}}
\authorrunning{J. Martinů, P. Šimánek}
%
\institute{Faculty of Information Technology, Czech Technical University in Prague, Prague 16000, Czech Republic}
\maketitle              
\begin{abstract}

This study investigates the application of deep-learning diffusion models for the super-resolution of weather data, a novel approach aimed at enhancing the spatial resolution and detail of meteorological variables. Leveraging the capabilities of diffusion models, specifically the SR3 and ResDiff architectures, we present a methodology for transforming low-resolution weather data into high-resolution outputs. Our experiments, conducted using the WeatherBench dataset, focus on the super-resolution of the two-meter temperature variable, demonstrating the models' ability to generate detailed and accurate weather maps. The results indicate that the ResDiff model, further improved by incorporating physics-based modifications, significantly outperforms traditional SR3 methods in terms of Mean Squared Error (MSE), Structural Similarity Index (SSIM), and Peak Signal-to-Noise Ratio (PSNR). This research highlights the potential of diffusion models in meteorological applications, offering insights into their effectiveness, challenges, and prospects for future advancements in weather prediction and climate analysis.

\keywords{Weather modelling  \and Super-resolution \and Denoising diffusion probabilistic models.}
\end{abstract}
\section{Introduction}
The application of deep learning models, particularly diffusion models, to enhance the resolution of weather data represents a significant advancement in the field of artificial intelligence and meteorology. This research area is driven by the critical need to improve the accuracy and detail of weather predictions, which are essential for a wide range of applications, from agriculture and aviation to disaster management and climate research. Super-resolution (SR) techniques, traditionally applied to image processing, are being adapted to meteorological data to generate high-resolution (HR) outputs from low-resolution (LR) inputs, thereby providing more detailed and accurate weather information.

Diffusion models, a class of generative models that have shown remarkable success in generating high-quality images, offer a promising approach to super-resolution. These models work by gradually denoising a signal, starting from a random distribution and moving towards the data distribution, effectively refining the details of images in a controlled manner. In the context of weather data, this technique can be used to enhance the resolution of various meteorological variables, such as temperature, precipitation, and wind speed maps, which are crucial for accurate weather forecasting and climate analysis.

The motivation behind using deep-learning diffusion models for the super-resolution of weather data lies in their ability to capture complex atmospheric patterns and details that are often missed or smoothed out in lower-resolution datasets. By generating higher-resolution weather data, we can achieve more precise local weather predictions, improve the understanding of microclimates, and enhance climate models. This not only benefits scientific research but also has practical implications for agriculture, urban planning, and emergency response strategies, where detailed weather information can lead to better decision-making and outcomes.

The application of such advanced super-resolution techniques to weather data is still an emerging field, with significant research potential. This paper aims to contribute to this field by exploring the use of deep-learning diffusion models for the super-resolution of weather data, focusing on their effectiveness, challenges, and potential improvements. By doing so, we seek to open new pathways for enhancing the quality and utility of meteorological data, contributing to more accurate and detailed weather forecasts and climate models.

\section{Related Literature}

The literature on the super-resolution of weather data using deep learning, particularly diffusion models, intersects with several research domains: image super-resolution, generative models, and meteorological data enhancement.

Image Super-Resolution with Deep Learning: The foundation of using deep learning for super-resolution is well-established, with Convolutional Neural Networks (CNNs) being the most common approach. Techniques like SRCNN \cite{dong2015image} and ESRGAN \cite{wang2018esrgan} have shown significant improvements in generating high-resolution images from low-resolution counterparts. These methods have set the stage for applying deep learning to various SR tasks, including weather data.

Diffusion Models in Generative Tasks: Diffusion models, such as those proposed by \cite{ho2020denoising_DDPM} and further developed by \cite{saharia2021image_SRDDPM}, represent a newer class of generative models that have demonstrated remarkable capabilities in generating high-quality images. These models operate by reversing a diffusion process, gradually denoising an image from a purely noisy state to a detailed high-resolution image. Their application to image super-resolution, termed SR3 by \cite{saharia2021image_SRDDPM}, has shown promising results in enhancing image details while maintaining natural textures and patterns.

Super-Resolution of Meteorological Data: While the application of super-resolution techniques to meteorological data is less explored, there is a growing interest in this area. Studies like those by \cite{vandal2017deepsd} and \cite{2019GL082532} have applied machine learning to downscale climate and weather models, improving the spatial resolution of precipitation forecasts and cloud structures. These approaches highlight the potential of machine learning in enhancing the resolution and accuracy of weather predictions but also point to the need for specialized models that can handle the unique challenges of meteorological data, such as its spatiotemporal dynamics and non-linear relationships.

Challenges and Opportunities: The adaptation of diffusion models to weather data super-resolution presents both challenges and opportunities. One challenge is the need to model the complex and dynamic nature of weather patterns accurately. However, the inherent flexibility and capability of diffusion models to capture intricate details offer a unique opportunity to significantly improve the resolution and quality of weather data.

In conclusion, the literature indicates a promising intersection of deep learning, diffusion models, and meteorological data enhancement. This paper builds on these foundations, aiming to advance the field by specifically focusing on the application of deep-learning diffusion models for the super-resolution of weather data, addressing both the challenges and potential of this innovative approach.

\section{Methods}
\subsection{SR3}
SR3, as first proposed by \cite{saharia2021image_SRDDPM}, is a model developed for image super-resolution (SR) that incorporates diffusion models \cite{ho2020denoising_DDPM}. It enhances low-resolution images by iteratively refining their details and quality, aiming for high-resolution outputs.

Diffusion models are characterized by two primary processes described through Markov chains: the forward process and the reverse process. The forward process progressively adds Gaussian noise to an initial image \(y_0\), following a predefined noise schedule \((\beta_1, \beta_2,...,\beta_T)\). For each step \(t\), let \(\alpha_t = 1 - \beta_t\) and \(\bar{\alpha}_t = \prod_{i=1}^{t} \alpha_i\). The distribution of the noised image \(y_t\) given the original image \(y_0\) at step \(t\) can then be expressed as:  
$$
q\left(y_t \mid y_0\right) = \mathcal{N}\left(y_t ; \sqrt{\bar{\alpha}_t} y_0, (1-\bar{\alpha}_t) I\right)
$$
where 

In the reverse process, the goal is to reconstruct the original image \(y_0\) from the noise, where the distribution of noise is \(\mathcal{N}(\mathbf{0}, \boldsymbol{I})\). Direct computation of \(q(x_t|x_{t-1})\) is impossible; instead, reconstruction is approximated by a learned model \(p_\theta(x_{t-1}|x_t)\) with parameters \(\theta\) where  \(x\) is low resolution source image:
$$\displaystyle{p}_{{\theta}}{\left({y}_{{{t}-{1}}} \mid {y}_{{{t}}},{x}\right)}={\mathcal{N}}{\left({y}_{{{t}-{1}}} ; \mu_{{\theta}}{\left({x},{y}_{{{t}}},\bar{\alpha}_t\right)},{\sigma_{{{t}}}^{{{2}}}}{I}\right)}$$

Training involves simulating both the forward and reverse processes on training images, and optimizing the model to maximize the likelihood of reconstructing the high-resolution (HR) target image. This optimization targets minimizing the negative log-likelihood, which, after applying Jensen's Inequality and a series of derivations including reparametrization, leads to the following objective function:
$$
\mathbb{E}_{t, y_0, x, \mathbf{e}_t}\left[\left\|\mathbf{e}_t - f_\theta\left(x,\sqrt{\bar{\alpha}_t} y_0 + \sqrt{1-\bar{\alpha}_t} \mathbf{e}_t, t\right)\right\|^2\right]
$$
Here, \(f_\theta\) is the model predicting the noise added at step \(t\).

Inference with SR3 begins by sampling a noisy image \(y_t \sim \mathcal{N}(\mathbf{0}, \boldsymbol{I})\) and iteratively denoising it towards a high-resolution output. At each step \(t\), noise \(z \sim \mathcal{N}(\mathbf{0}, \boldsymbol{I})\) is sampled and used to update the image according to:
$$\displaystyle{y}_{{{t}-{1}}}=\frac{{{1}}}{{\sqrt{{\alpha_{{{t}}}}}}}{\left({y}_{{{t}}}-\frac{{{1}-\alpha_{{{t}}}}}{{\sqrt{{{1}-\bar{\alpha}_t}}}}{{f}_{{\theta}}{\left({x},{y}_{{{t}}},\bar{\alpha}_t\right)}}\right)}+\sqrt{{{1}-\alpha_{{{t}}}}}{z}
$$
This iterative process is carried out for steps \(T, ..., 1\), ultimately yielding the denoised, high-resolution image.

The denoising function's \(f_\theta\) architecture, employs a U-net structure. The input to this function consists of an interpolated image concatenated with a noisy image. The U-net incorporates ResNet blocks \cite{resnet_block} with self-attention and skip connections. A more comprehensive description of the architecture can be found in \cite{saharia2021image_SRDDPM} and in Figure 1, where an enhanced version of the SR3 architecture, resdiff \cite{resdiff}, is presented. All the other details regarding SR3 are provided in \cite{saharia2021image_SRDDPM}.

\subsection{ResDiff}

\begin{figure}[h]
    \centering
    \includegraphics[width=1.1\columnwidth]{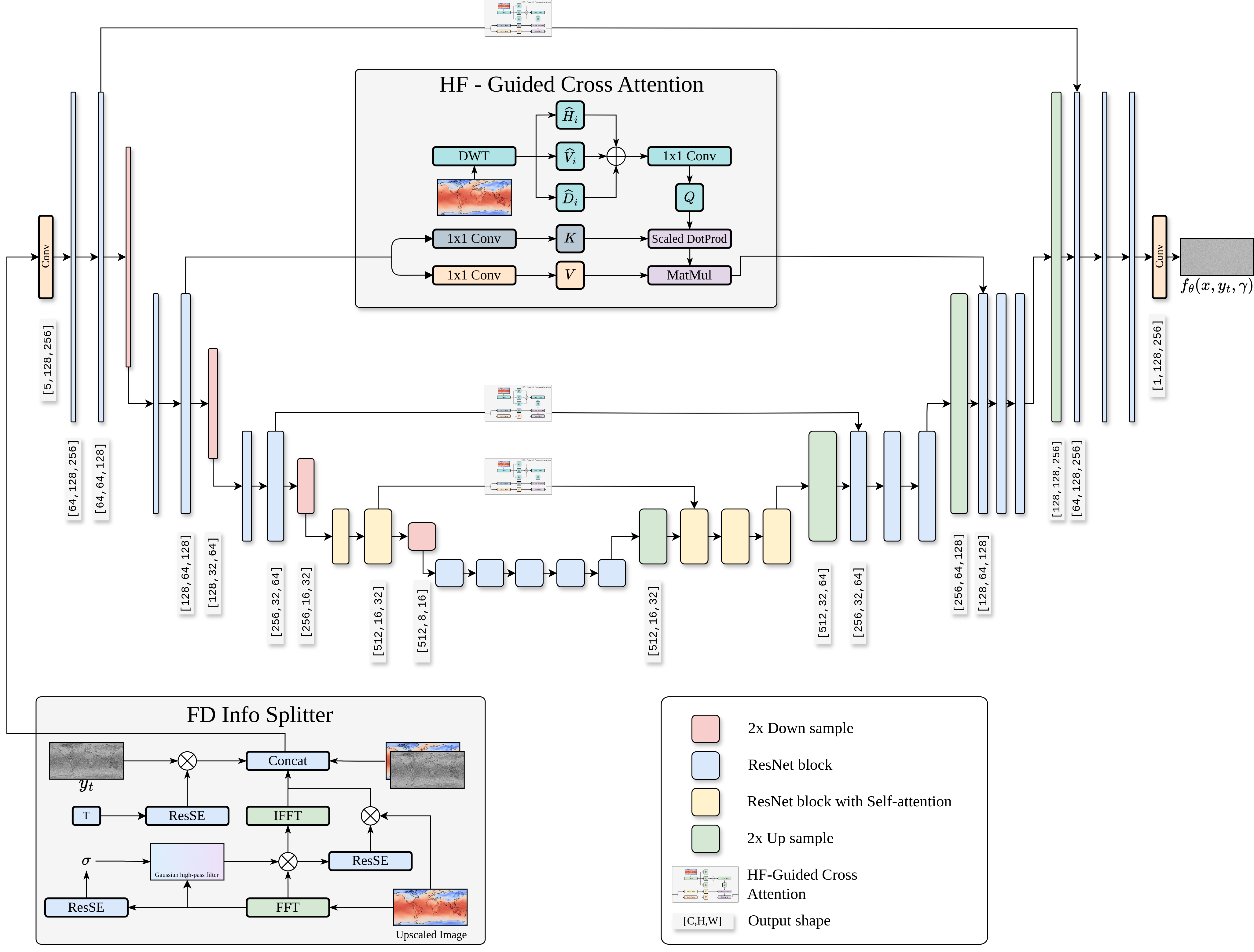}
    \caption[resdiff architecture]{Architecture of Resdiff model used for Climate Variable downscaling. }
    \label{fig:resdiff_architecture}
\end{figure}

The Residual-structure-based diffusion model \cite{resdiff} is an architecture based on SR3 that combines Convolutional Neural Networks (CNNs) \cite{cnn_oshea2015introduction} and modified Diffusion Probabilistic Models.

ResDiff substitutes the initial bicubic interpolation prediction with a pre-trained CNN, specialized in capturing major low-frequency components and partial high-frequency components. However, we do not employ this pre-trained CNN because it performs less effectively with climatic data compared to classical bicubic interpolation. 

Furthermore, ResDiff introduces High-Frequency Guided Diffusion that modifies the diffusion model to focus on and enhance high-frequency image components. This method ensures that the details lost in the SR3 diffusion process, are meticulously reconstructed. As shown in Figure \ref{fig:resdiff_architecture}, HF Guided Diffusion adds the FD Info Splitter and HF-guided Cross-Attention into the architecture. 

\subsubsection{FD Info Splitter}
The Frequency-Domain Information Splitter (FD Info Splitter) \cite{resdiff} is a novel component of ResDiff that segregates image data into high and low-frequency bands. This segregation is fundamental for focusing the model's attention on preserving or enhancing the details critical for high-quality image super-resolution. The process begins with the application of a 2D Fast Fourier Transform (FFT) to both the interpolated and the noised images. This transformation allows the model to analyze and manipulate the frequency components of the image data directly.

Another innovation within the FD Info Splitter is the use of a Residual Squeeze-and-Excitation (ResSE) block, proposed and described in \cite{ResSe}. The ResSE block processes the FFT-transformed feature maps, dynamically adjusting the importance of different channels in the output feature map. Following this, the computation of the standard deviation, \(\sigma\), is carried out with formula:
$$
\sigma = \min \left(|\operatorname{ResSE}(M)| + \frac{l}{2}, l\right)
$$
where \(M\) represents the feature map obtained from the FFT, and \(l\) denotes a predetermined upper limit for the filter's standard deviation. This calculation ensures that the emphasis on high-frequency details is adaptively scaled, enhancing the model's focus on areas of the image requiring finer detail reconstruction. The high-pass filtering, represented by \(H(u, v)\), is then applied to isolate and enhance these high-frequency components:
$$
H(u, v) = 1 - e^{-\frac{D^2(u, v)}{2 \sigma^2}}
$$
where \(D(u, v)\) measures the distance of a frequency component from the origin in the frequency domain. This filtering process results in a feature map \(L\), enriched with high-frequency details. \cite{resdiff}

Following the high-pass filtering, an inverse FFT is applied to \(L\), facilitating the generation of a low-frequency focused image representation, \(x_{LF}\). Concurrently, the ResSE block refines \(L\) to extract attention weights that are specific to the high-frequency domain. These weights are then applied to the interpolated image to isolate its high-frequency components, resulting in \(x_{HF}\).

The culmination of this process produces a set of five distinct feature maps: \([x, y, x_{HF}, x_{LF}, x'_{t}]\), where \(x\) and \(y\) represent the source and the noise-added images, respectively, \(x_{HF}\) and \(x_{LF}\) denote the high and low-frequency components extracted, and \(x'_{t}\) is the target high-resolution output at iteration \(t\). These feature maps are integral to the ResDiff model's ability to reconstruct images with enhanced detail and clarity, particularly in the high-frequency bands that are crucial for perceptual quality \cite{resdiff}.

\subsubsection{HF-guided Cross-Attention}
The HF-guided Cross-Attention \cite{resdiff} mechanism further refines the model's capacity to focus on and enhance high-frequency details. By leveraging the high-frequency component obtained by Discrete wavelet transformation \cite{wavelets}, this mechanism directs the model's attention specifically towards areas of the image that benefit most from detail enhancement. This targeted approach ensures that the diffusion process is finely tuned to the nuances of image textures and edges, significantly improving the reconstruction of details lost in the initial SR3 diffusion stages.

In summary, ResDiff's architecture, with its emphasis on high-frequency detail preservation through innovative mechanisms like the FD Info Splitter and HF-guided Cross-Attention, represents a significant advancement in the field of image super-resolution. It addresses the limitations of previous models by providing a more focused and effective method for reconstructing images with high fidelity, especially in applications sensitive to the loss of high-frequency information.

\begin{figure}[h]
    \includegraphics[width=1\columnwidth]{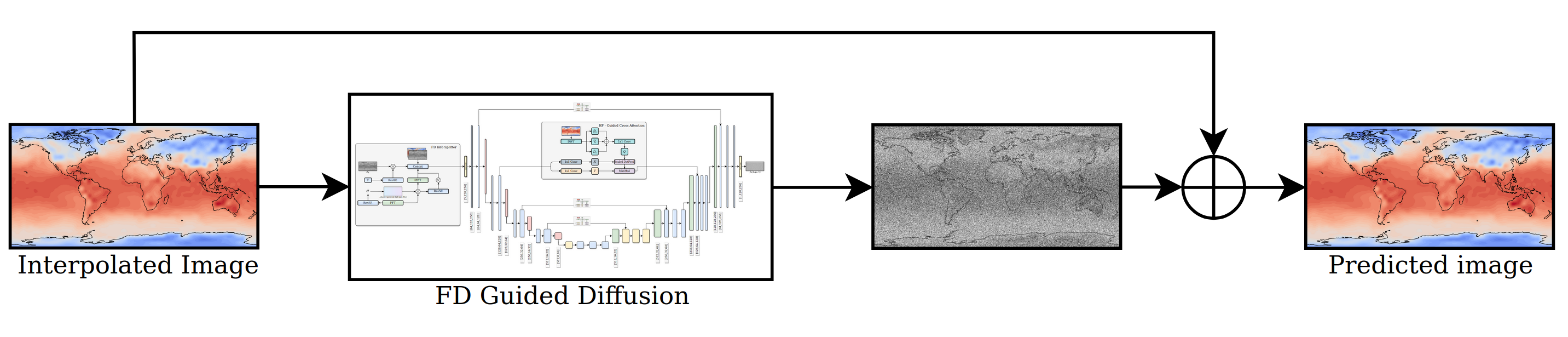}
    \caption[resdiff architecture]{Resdiff image sampling architecture.}
    \label{fig:resdiff_img_concat}
\end{figure}

It's significant to mention that, the ResDiff architecture diverges from SR3 by adding the image outputted by the U-net to the bicubic-interpolated image, as depicted in Figure \ref{fig:resdiff_img_concat}.

\subsection{ResDiff Enhanced with Physics-Inspired Convolutional Filters}

In an effort to further refine the ResDiff architecture for applications requiring nuanced understanding of dynamic systems, such as weather prediction, we introduce a novel modification termed "ResDiff + Physics". This adaptation integrates physics-inspired convolutional filters that mimic finite difference schemes used for computing derivatives. This approach allows the model to explicitly focus on derivative features that are fundamental to the Navier-Stokes equations, which govern fluid dynamics and are critical in meteorological modeling.

\paragraph{Motivation}
The integration of physics-based principles into deep learning models offers a promising direction for improving the accuracy of predictions in fields heavily reliant on physical laws. By embedding convolutional filters that approximate spatial derivatives, "ResDiff + Physics" aims to capture the underlying physical processes that drive weather patterns. This method not only enhances the model's capacity for detail and pattern recognition but also aligns its internal feature extraction mechanisms more closely with the real-world phenomena it seeks to emulate.

\paragraph{Convolutional Derivative Filters}
To achieve this, we replace the Frequency-Domain Information Splitter with a set of three specialized convolutional filters designed to approximate first and second-order spatial derivatives, key components in differential equations like the Navier-Stokes. These filters are applied to the interpolated image, capturing the gradient and curvature information relevant to fluid motion and atmospheric dynamics. The filters are defined as follows:

\begin{minipage}{.3\textwidth}
\[
\text{} \partial_x = 
\begin{bmatrix}
0 & 0 & 0 \\
0 & -1 & 1 \\
0 & 0 & 0 \\
\end{bmatrix}
\]
\end{minipage}
\begin{minipage}{.3\textwidth}
\[
\partial_y = 
\begin{bmatrix}
0 & 0 & 0 \\
0 &  -1 &  0 \\
0 &  1 &  0 \\
\end{bmatrix}
\]
\end{minipage}
\begin{minipage}{.3\textwidth}
\[
\nabla^2 = 
\begin{bmatrix}
0 & 1 & 0 \\
1 & -4 & 1 \\
0 & 1 & 0 \\
\end{bmatrix}
\]
\end{minipage}      
\vspace{10pt}

Each filter is applied to the interpolated image using a reflect padding of 1 to ensure boundary consistency. The resulting derivative feature maps are then concatenated, forming a comprehensive tensor of shape \([B, 3, 128, 256]\) that encapsulates spatial variation information.

\paragraph{Integration into ResDiff}
Following the convolutional operation, the derivative feature maps are concatenated with the noise-added image and the original interpolated image. This enriched tensor serves as the input to the U-net architecture, conditioning the model on physically meaningful derivatives rather than solely on frequency-domain information. This modification aims to ground the model's learning process in the physical reality of atmospheric dynamics, providing a more informed basis for generating high-resolution weather predictions.

Furthermore, we have refined the HF-guided cross-attention mechanism by incorporating a 1x1 convolution directly applied to the high-frequency components derived from the Discrete Wavelet Transform (DWT). This adjustment further enhances the model's ability to focus on and reconstruct high-frequency details, now informed by derivative-based features that reflect underlying physical processes.

"ResDiff + Physics" represents the first advancement in the application of deep learning to weather prediction and other fields governed by complex physical laws. By integrating physics-inspired convolutional filters, the model is well-equipped to interpret and reconstruct images in a manner that is both high-fidelity and physically coherent, promising improvements in the accuracy and reliability of predictions for dynamic systems.

\section{Experiments}
\subsection{Data}
For model training and evaluation, we used the WeatherBench dataset \cite{Rasp_2020_weatherbenchdataset}, a benchmark dataset designed to evaluate and compare machine learning models on weather forecasting tasks. It includes a range of meteorological variables derived from the ERA5 \cite{era5} reanalysis dataset, which is a comprehensive dataset combining model data with observations from across the world to provide a consistent, gridded view of the weather at various points in the past.

In this work, our focus is on the downscaling of the T2M (two-meter temperature) variable.
This variable represents temperature readings in Kelvin, mapped across a latitude-longitude grid situated two meters above the ground, with data recorded hourly. Consequently, each day is represented by 24 distinct temperature measurements,
Since our work is a super-resolution task, we used data pairs with different grid spacings of 5.625° and 1.40525°. As a result, the low-resolution image consists of 32 × 64 pixels, while the high-resolution image contains 128 × 256 pixels. These images consist of only one channel as we are working with a single variable.
For model training, we used data pairs collected between January 1, 1979, and February 1, 2015. The validation set covers the period from January 1, 2016, to February  1, 2016. 
Due to limited computational resources and the slow validation of our model when utilizing 1000 timesteps for sampling, we downsized the dataset by using only the values measured in January for both the validation and training sets. These data are then standardized separately for each resolution.

\subsection{Description of the experiments}
All experiments were conducted on a single NVIDIA A100 graphics card. We trained models for 200,000 iterations, using various batch sizes. Implementation of the code was done in PyTorch.
Specifically, SR3 models utilized a batch size of 16, while ResDiff required a batch size of 4 due to its higher memory demands during experimentation.
Validation was performed every 10,000 iterations across the entire validation set. 
Despite the computational intensity of diffusion models during inference, as they must utilize the entire U-net for each timestep T, we kept T=1000 for both training and validation to have accurate validation error estimates. As a result, the total training time for a single run was approximately 50 hours. We used a linear noise schedule from 1e-6 to 1e-2 and a dropout rate of 0.2. Additionally, similar to the authors in the original paper \cite{ho2020denoising_DDPM}, we applied Exponential Moving Average to model parameters with a decay rate of 0.9999. For optimization, we utilized Adam with a learning rate of 1e-4.
Overall, we aimed to maintain the hyperparameters close to those of the original ResDiff \cite{resdiff} model.

As validation metrics, we used MSE and MAE for error analysis, along with SSIM, which assesses image quality by structural similarity, and PSNR, measuring the signal-to-noise ratio to gauge image reconstruction precision. \cite{psnr}

\section{Results}
\begin{figure}[h]
    \includegraphics[width=1\columnwidth]{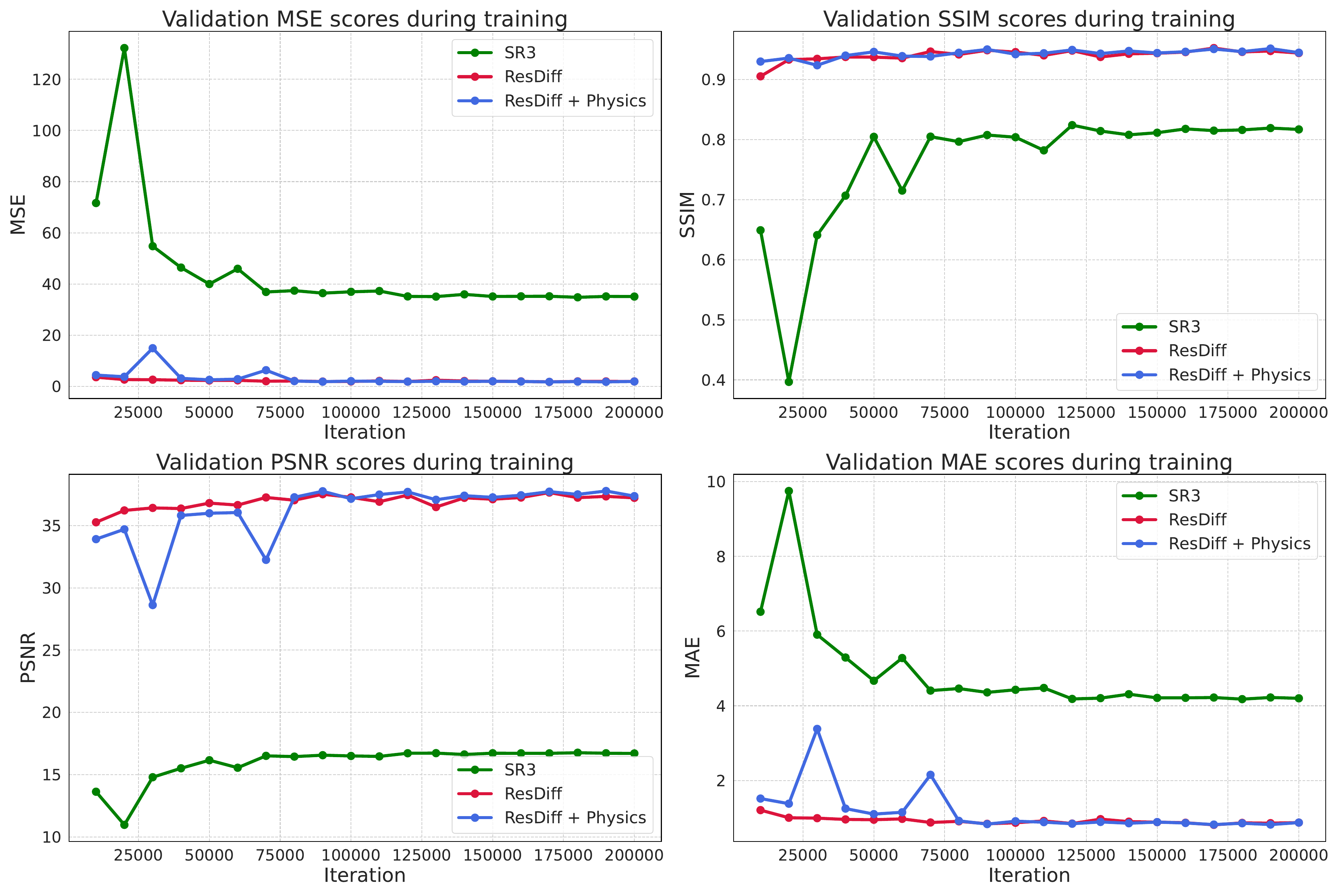}
    \caption[Validation scores]{Validation scores during training across models: SR3, ResDiff, ResDiff + Physics.}
    \label{fig:validated_training_run}
\end{figure}

Using SR3 as a benchmark we can observe from Figure \ref{fig:validated_training_run}, that both ResDiff and ResDiff+Physics significantly outperform the SR3 method throughout the entire 200,000 iterations. The validation accuracy of SR3 is not stable and exhibits considerable fluctuation during training. In contrast, ResDiff-based methods maintain an SSIM above 0.9 consistently. SR3 achieves its best validation SSIM at 120,000 iterations, while ResDiff+Physics and ResDiff reach their peak validation SSIM at 190,000 and 170,000 iterations, respectively.

\begin{figure}[H]
    \includegraphics[width=0.96\columnwidth]{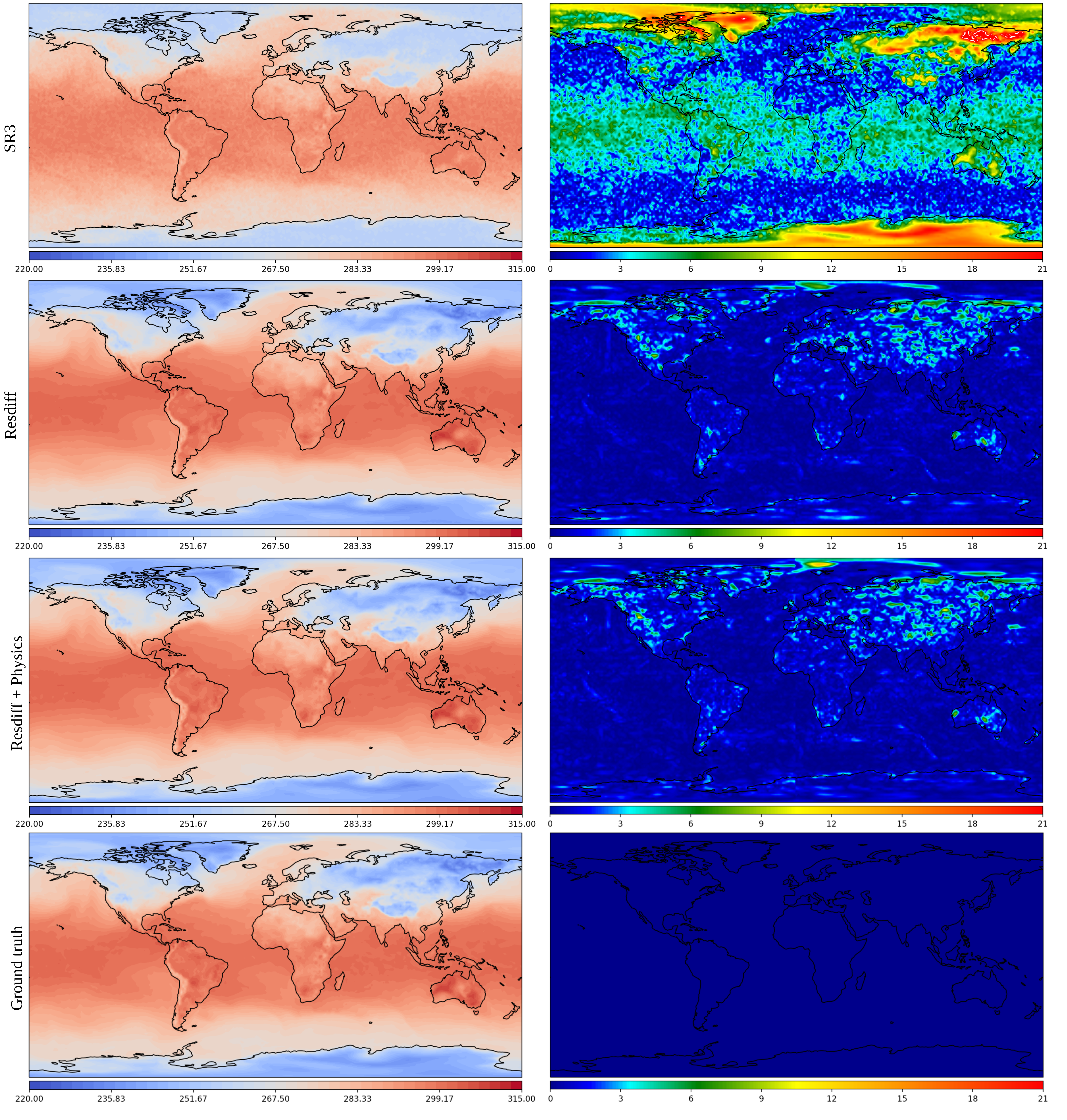}
    \caption[Comparison of images generated]{In the left column, the HR reference image and images generated by the models are displayed, annotated with their corresponding temperatures in Kelvin. The right column shows the absolute error between HR reference and super-resolution images.}
    \label{fig:generated_images}
\end{figure}
Overall, as shown in Table \ref{tab:validation_scores}, the ResDiff model achieves the best results, except for PSNR, where it is outperformed by the ResDiff+Physics architecture. In contrast, the SR3 model shows a much higher MSE.

\begin{table}[H]
\centering
\caption{Best validation scores across all models.}
\setlength{\tabcolsep}{12pt} 
\begin{tabular}{lccccc}
\hline
Method & MSE \(\downarrow\) & SSIM \(\uparrow\) & PSNR \(\uparrow\) & MAE \(\downarrow\) \\ \hline
Ground Truth & 0 & 1 & \(\infty\) & 0 \\ \hdashline
SR3 & 35.15 & 0.824 & 16.71 & 4.477 \\
ResDiff & \textbf{1.768} & \textbf{0.952} &  37.65  & \textbf{0.813} \\
ResDiff + Physics & 1.789 &  0.951 & \textbf{37.78}  & 0.821 \\ \hline
\end{tabular}
\label{tab:validation_scores}
\end{table}
The outputs from SR3 are worse than interpolated outputs. On the other hand, ResDiff and ResDiff+Physics capture low-frequency information much more effectively, yielding results that are difficult to distinguish from high-resolution reference images, as illustrated in Figure \ref{fig:generated_images}.

\section{Discussion}

The findings from this study underscore the potential of deep-learning diffusion models, particularly SR3 and its enhancements via the ResDiff and ResDiff+Physics architectures, in the super-resolution of weather data. These models have demonstrated the ability to significantly improve the resolution and detail of meteorological variables, offering promising avenues for enhancing weather prediction accuracy and climate model fidelity.

The comparison of SR3 with ResDiff and ResDiff+Physics highlights the importance of model architecture and training strategy in achieving high-quality super-resolution. The superior performance of ResDiff and ResDiff+Physics over SR3, as evidenced by the validation metrics, can be attributed to their tailored design for capturing and enhancing high-frequency details, which are crucial in representing the fine-scale features of weather patterns. This suggests that the integration of domain-specific knowledge, such as the physics of weather phenomena, into the diffusion model architecture can further improve super-resolution outcomes.

Moreover, the application of these models to the WeatherBench dataset — a benchmark dataset for evaluating machine learning models on weather forecasting tasks — provides a concrete example of how advanced super-resolution techniques can be applied to real-world meteorological data. The ability of ResDiff and ResDiff+Physics to outperform traditional SR methods, including SR3, in terms of metrics such as MSE, SSIM, and PSNR, demonstrates their potential to contribute meaningfully to the fields of meteorology and climate science.

\subsection{Implications and Future Work}

The implications of this research extend beyond the academic domain, offering valuable insights for practical applications in weather forecasting, climate modeling, and environmental monitoring. By improving the resolution and accuracy of weather data, these models can aid in more precise local weather predictions, enhance the understanding of climate change impacts, and support decision-making in sectors such as agriculture, urban planning, and disaster management.

Future work should focus on further refining diffusion model architectures and training strategies to enhance their efficiency and effectiveness in super-resolving weather data. This includes exploring the integration of additional physical constraints and variables into the models to ensure that the super-resolved data adheres closely to real-world atmospheric dynamics. Additionally, expanding the application of these models to a broader range of meteorological variables and datasets will be crucial in assessing their generalizability and utility in diverse climatic and geographical contexts.

\section{Conclusion}

This study has demonstrated the effectiveness of deep-learning diffusion models, specifically SR3, ResDiff, and ResDiff+Physics, in the super-resolution of weather data. By leveraging the capabilities of these models to capture and enhance fine-scale details in meteorological variables, we have shown that it is possible to significantly improve the resolution and quality of weather data, thereby contributing to more accurate and detailed weather forecasts and climate models. The success of these models underscores the value of interdisciplinary research, blending advances in artificial intelligence with meteorological science, to address some of the most pressing challenges in weather prediction and climate analysis. As we continue to refine these models and explore their applications, we move closer to realizing the full potential of deep learning in enhancing our understanding and prediction of weather and climate phenomena.

\bibliographystyle{splncs04}
\bibliography{bib}

\end{document}